\documentclass[letterpaper, 10 pt, conference]{ieeeconf}  

\usepackage{amsmath,amssymb}
\usepackage[overload]{empheq}
\usepackage{graphicx}
\usepackage[export]{adjustbox}
\usepackage[dvips]{epsfig}
\usepackage{epstopdf}
\usepackage[table]{xcolor}
\usepackage{url}
\usepackage{subfig}
\usepackage{multirow}
\usepackage{caption}
\usepackage{tikz}

\newcommand\copyrighttext{%
  \centering\footnotesize \textbf{\copyright 2023 IEEE}.  Personal use of this material is permitted.  Permission from IEEE must be obtained for all other uses, in any current or future media, including reprinting/republishing this material for advertising or promotional purposes, creating new collective works, for resale or redistribution to servers or lists, or reuse of any copyrighted component of this work in other works. \\
  }
\newcommand\copyrightnotice{%
\begin{tikzpicture}[remember picture,overlay]
\node[anchor=south,yshift=-2cm] at (current page.north) {\fbox{\parbox{\dimexpr0.9\textwidth-\fboxsep-\fboxrule\relax}{\copyrighttext}}};
\end{tikzpicture}%
}

\graphicspath{{./FIGURES/}} 

\makeatletter
\def\maketag@@@#1{\hbox{\m@th\normalfont\normalsize#1}}
\makeatother

\newcommand*\xbar[1]{%
  \hbox{%
    \vbox{%
      \hrule height 0.5pt 
      \kern0.5ex
      \hbox{%
        \kern-0.1em
        \ensuremath{#1}%
        \kern-0.1em
      }%
    }%
  }%
}

\makeatletter
\DeclareRobustCommand{\atan}{%
  \operatorname{atan}%
  \@ifnextchar2{_}{}%
}
\makeatother

\IEEEoverridecommandlockouts                              

\title{\LARGE \bf
Vehicle fuel consumption virtual sensing from GNSS and IMU measurements}

\author{Marcello Cellina, Silvia Strada and Sergio Matteo Savaresi$^{1}$
\thanks{$^{1}$Marcello Cellina, Silvia Strada, and Sergio Matteo Savaresi are with the Dipartimento di Elettronica, Informazione e Bioingegneria, Politecnico di Milano, Piazza Leonardo da Vinci 32, 20133 Milan, Italy. Email: 
\tt\small{marcello.cellina, silvia.strada, sergio.savaresi}@polimi.it}
}

\begin{document}

\maketitle

\copyrightnotice

\thispagestyle{empty}
\pagestyle{empty}

\begin{abstract}
This paper presents a vehicle-independent, non-intrusive, and light monitoring system for accurately measuring fuel consumption in road vehicles from longitudinal speed and acceleration derived continuously in time from GNSS and IMU sensors mounted inside the vehicle. In parallel to boosting the transition to zero-carbon cars, there is an increasing interest in low-cost instruments for precise measurement of the environmental impact of the many internal combustion engine vehicles still in circulation. The main contribution of this work is the design and comparison of two innovative black-box algorithms, one based on a reduced complexity physics modeling while the other relying on a feedforward neural network  for black-box fuel consumption estimation using only velocity and acceleration measurements. Based on suitable metrics, the developed algorithms outperform the state of the art best approach, both in the instantaneous and in the integral fuel consumption estimation, with errors smaller than 1\% with respect to the fuel flow ground truth. The data used for model identification, testing, and experimental validation is composed of GNSS velocity and IMU acceleration measurements collected during several trips using a diesel fuel vehicle on different roads, in different seasons, and with varying numbers of passengers. Compared to built-in vehicle monitoring systems, this methodology is not customized, uses off-the-shelf sensors, and is based on two simple algorithms that have been validated offline and could be easily implemented in a real-time environment.

\end{abstract}

\section{INTRODUCTION} \label{sec:intro}
Transportation worldwide accounts today for $17\%$ of global greenhouse gas emissions, \cite{statista17global}, behind only the power sector. Road vehicles are the leading source of transportation emissions, and passenger cars alone accounted for $41\%$ of global transportation-related CO2 emissions in 2020, \cite{statista41transport}. 
As a result, apart of course from boosting the transition to low and zero-carbon vehicles, there is a growing interest in improving the fuel efficiency of circulating Internal Combustion Engine Vehicles (ICEV) in order to minimize their environmental impact. One way to achieve this is by accurately estimating vehicle fuel (diesel or petrol) consumption, which can help identify areas for improvement both in vehicle design and road traffic regulation.

Global Navigation Satellite Systems (GNSS) and Inertial Measurement Unit (IMU) sensors are cheap, abundant, easy to install, and do not require any access to vehicle control electronics. A vehicle fuel consumption estimator based only on velocity and acceleration measurements provided by these sensors, appropriately installed on the means of transport, could be incorporated in a wide range of vehicles, as it does not require specialized hardware and thus allows the estimation to be vehicle-independent. 

In order to correctly estimate an ICEV fuel consumption, it is necessary to model the correlation between the fuel flow dynamics and the vehicle motion. Historically, a number of models have been proposed in the literature. These models have been investigated in a number of survey articles, for example, \cite{faris2011vehicle} and  \cite{zhou2016review}.

White box physical models, as in \cite{heywook1988internal} and in \cite{an1997development}, employ an accurate model of the main components of an ICEV vehicle motion and engine efficiency, as in \cite{hellstrom2009look}. These very detailed models, although useful in the design phase of an engine or for solving optimization problems, require in-depth knowledge of the design parameters of the  vehicle, which are not publicly available most of the time. 

Some grey-box approaches have been studied based on the knowledge of the instantaneous engine power demand, as in \cite{post1984fuel} and \cite{leung2000modelling}. These models are useful for the characterization of an ICEV if a dynamo-metric test bench is available to precisely measure the instantaneous engine power, otherwise they need to be complemented with a model of the resistive forces of the vehicle, as in \cite{wang2017fuel}.
Another grey-box approach, presented in \cite{saerens2013assessment}, uses a polynomial model combining the engine's rotational speed and an additional input that can be the engine's instantaneous power, torque, or load factor, giving in output the instantaneous ICEV fuel consumption. The limitation of this approach is that it relies on the measurement of quantities that are difficult to obtain without direct access to the ICEV control unit.

A black-box approach is presented in \cite{hung2005modal}. This model divides the driving phase into four different modes: idle, acceleration, cruise, and deceleration. For each of these modes, a relationship between the velocity and the fuel consumption is established, aside from the first, where the relationship is between fuel consumption and time. However, since the acceleration and road slope dynamics are neglected, the model exhibits low predictive accuracy.
The most important class of models related to our application refers to black-box models that estimate the instantaneous fuel consumption using velocity and acceleration measurements of the car body. The most frequently cited model in literature belonging to this category is the VT-MICRO model originally presented in \cite{ahn2002estimating} and then expanded in \cite{park2010development} and \cite{lei2010microscopic}.
The VT-MICRO model is considered state-of-the-art concerning the problem of estimating the instantaneous fuel flow of an ICEV using only velocity and acceleration measurements. However, this model does not work properly outside its training envelope, and it does not take into account phenomena such as ECU fuel cutoff during coasting down driving.

Some approaches use Neural Networks to solve this problem, like in \cite{kanarachos2019instantaneous} or in \cite{abediasl2023real}, although the low sampling rate and necessity of accessing additional vehicle and sensor data make them unsuitable for our applications.

In this work, we present and compare two innovative methodologies for black-box ICEV fuel consumption estimation using only GNSS and IMU measurements that outperform the state of the art, together with a set of metrics to evaluate their prediction performance and experimental results that demonstrate the superiority of our proposed methods over the existing approaches.

The structure of the paper is the following: in Section \ref{sec:set_up} are described the considered problem and the experimental setup. Then, Section \ref{sec:methods} outlines the proposed models and the model fitting methodologies, together with the comparison with the state-of-the-art VT-MICRO model approach. Finally, Section \ref{sec:results} illustrates the real dataset and the experimental results, highlighting the metrics employed to compare the instantaneous and integral consumption estimation performance.

\section{PROBLEM STATEMENT AND EXPERIMENTAL SETUP} \label{sec:set_up}

In this paper, a set of models for the estimation of vehicle fuel consumption using velocity and acceleration measurements is presented.
More specifically, given a synchronized pair of vehicle longitudinal velocity $v_x$ and longitudinal acceleration $a_x$ measurements, the model should provide an estimated fuel flow $f_f$ (hereafter, subscripts $_x$ and $_f$  will be omitted)  as:

\begin{equation}
    \hat{f_f} = \mathbf{F}(a_x, v_x)
\end{equation}

Note that there is no dependence on time, as the model is static and time-invariant.

The experimental setup is composed of a test vehicle and a combined GNSS/IMU sensing unit.

The vehicle used for this work is an Alfa Romeo Giulia with a 2.2l Diesel engine coupled with a 7-speed automatic gearbox. A measurement of the ground truth instantaneous fuel flow is published over the vehicle CAN network at a frequency of 20 Hz, with a resolution of 0.001$l/h$. The vehicle is equipped with a start-and-stop system which, if the battery is sufficiently charged, automatically turns off the engine when the vehicle is stopped.

The combined GNSS/IMU unit is a Suchy Data Systems XPro Nano 10, providing a measure of the vehicle's horizontal speed at a frequency of 10 Hz with a resolution of 0.01 $km/h$ and a 3-axis acceleration measure, aligned with the vehicle reference system, at a frequency of 100 Hz with a resolution of 0.001 $m/s^2$. 
The GNSS antenna is attached magnetically to the roof of the vehicle, while the combined IMU and GNSS receiver unit is mounted rigidly under the passenger's seat inside a custom 3D-printed support.

\begin{figure}
    \centering
    \includegraphics[width=\linewidth]{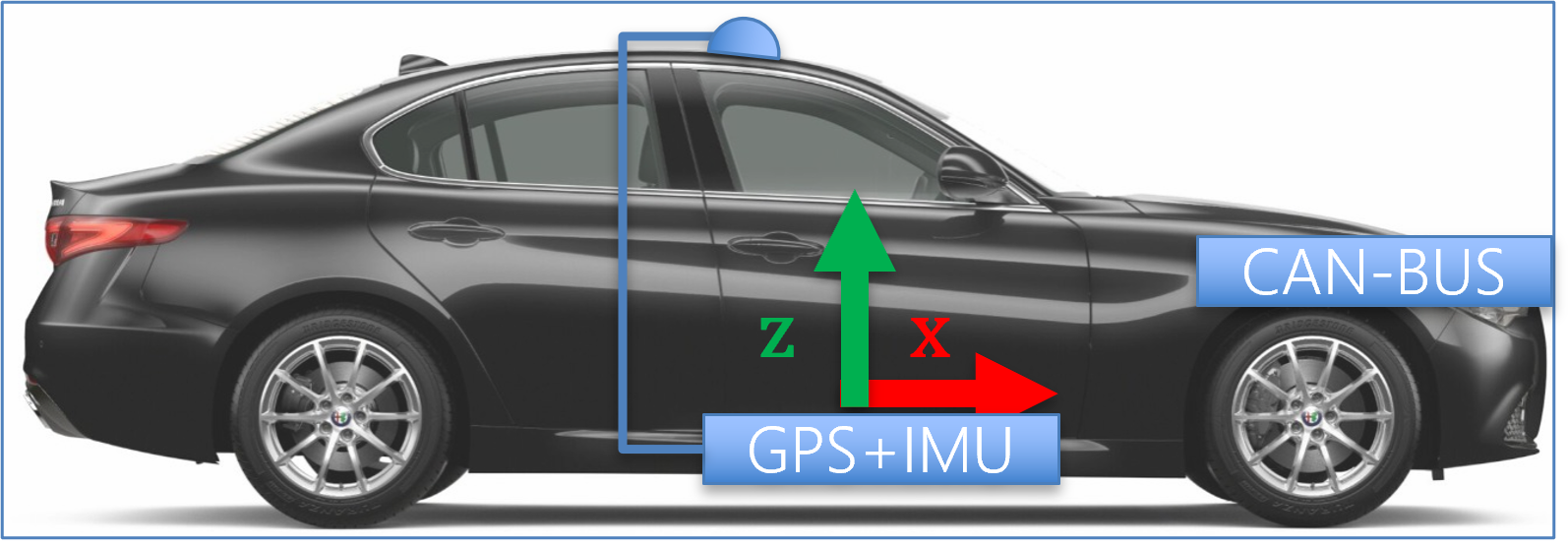}
    \caption{Experimental Setup}
    \label{fig:setup}
\end{figure}

The considered problem inputs and output are a set of synchronized $\{v,a,f\}$ measurements. Since the three signals are sampled at different frequencies, it is necessary to under-sample the higher frequency signals $a$ and $f$ to the lowest sampling frequency of the three (10 Hz for $v$ measurements). 

Note that under-sampling a signal could lead to aliasing if the raw signal has a significant frequency component outside the new Nyquist frequency, in this case, equal to 5 Hz or half of the GNSS measurements sampling frequency. 
Therefore, a zero-phase low-pass filter, which does not introduce any delay in the filtered data, is applied to the raw measurements of the three signals $v$, $a$, and $f$. The filter transfer function presents unitary gain and two coincident poles at a frequency of 3 Hz. 
Finally, a cubic spline is used to interpolate over the time domain the higher-frequency $a$ and $f$ signals to synchronize them with the exact sampling time of the velocity measurements.

\section{METHODOLOGIES} \label{sec:methods}

\subsection{State of the Art Black-box Model}

As mentioned in Section \ref{sec:intro}, the state-of-the-art solution to this problem is represented by the VT-MICRO black-box model, proposed in \cite{ahn2002estimating} and described by Equation \eqref{eq: vt_micro}:

\begin{equation} \label{eq: vt_micro}
    \log(\hat{f}) = \begin{dcases}
         \sum_{i = 0}^3\sum_{j = 0}^3L_{i,j} v^i a^j &\mbox{if } a \geq 0 \\
         \sum_{i = 0}^3\sum_{j = 0}^3M_{i,j} v^i a^j &\mbox{if } a < 0 
    \end{dcases}
\end{equation}

with $L_{i,j}$ being the model coefficients for the positive acceleration phase and $M_{i,j}$ being the model coefficients for the negative acceleration one. 
The fitting of this model is performed separately on positive and negative acceleration sub-sets of the total data, and the logarithmic transformation forces the estimated fuel flow to be positive. This model has a total of 32 coefficients to be identified and allows fitting using the linear least squares method on the unknown $\log(f)$. Note that this excludes all the data samples with $f=0$ from being used to train the model.

\subsection{Physics-Based (PM) Model}
\begin{figure}
    \centering
    \includegraphics[width=\linewidth]{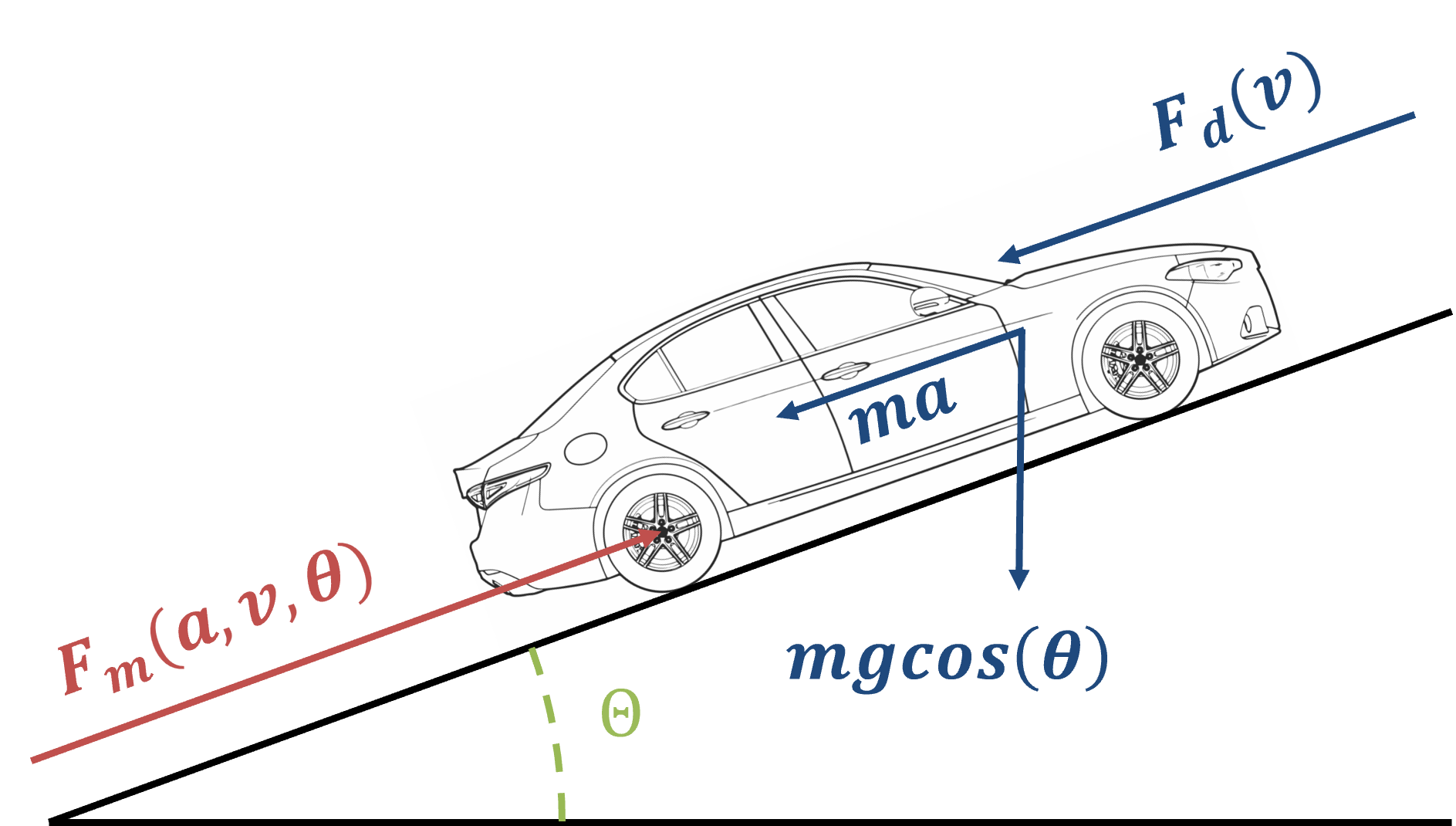}
    \caption{Longitudinal Model}
    \label{fig:mod-long}
\end{figure}

The first proposed model is a Physics-based model that comes from the longitudinal model of a vehicle driving on a terrain with slope $\theta$ at a velocity $v$ and acceleration $a$, as shown in Fig. \ref{fig:mod-long}.
The car traction force $F_m$ is balanced by three terms:
\begin{enumerate}
    \item The inertia of the vehicle mass, equal to $ma_x$
    \item The longitudinal component of the gravity force, equal to $mg\cos(\theta)$
    \item The ensemble of aerodynamic and rolling friction forces $F_d$
\end{enumerate}

The last term $F_d$ depends on the vehicle's aerodynamic and tire friction coefficients, but it can be expressed without losing generality as a quadratic polynomial of the vehicle speed, as expressed in \cite{genta1997motor}. Therefore, we can write:

\begin{equation}
    \label{longFm}
    F_m = ma_x+mg\cos(\theta)+\alpha +\beta v + \gamma v^2
\end{equation}

which, given the definition of power necessary to motion $P_m = F_m v$, becomes:

\begin{equation}
    \label{longPm}
    P_m = ma_xv+mg\cos(\theta)v+\alpha v+\beta v^2 + \gamma v^3
\end{equation}
              
An IMU sensor measuring the vehicle's longitudinal accelerations would necessarily measure also the effect of gravity so that the measured acceleration turns out to be $a = a_x + g\cos(\theta)$

The instantaneous fuel flow $f$ can be supposed to be proportional to the instantaneous power necessary for motion when the engine is providing traction power, resulting in 
\begin{equation}
    f = kP_m = k(mav+\alpha v+ \beta v^2 + \gamma v^3)
\end{equation}

This balance is valid only if the engine provides positive power, as the fuel flow becomes null when the vehicle brakes or coasts down.
If we consider the vehicle mass as one of the parameters to be estimated, the Physics-Based black-box model becomes:

\begin{equation}
    \label{eq: modelPB}
    \hat f = \begin{cases}
    \alpha_1 a v + \alpha_2 v + \alpha_3 v^2 \alpha_4 v^3   &\mbox{if } \hat f \geq 0 \\
    0                                                       &\mbox{if } \hat f < 0
    \end{cases}
\end{equation}

This model has 4 parameters and can be fitted using linear least squares over the unknown $\hat{f}$. As in the previous model, only data samples with $f>0$ have been used in the model training phase, as null fuel consumption data is outside the model validity domain.

\subsection{Neural Network}
The second proposed model is a feed-forward Neural Network (NN). The network is composed of a single hidden layer, an input layer of size 2, as the regressor variables are $v$ and $a$, and a one-dimensional output layer.

The network is trained by minimizing the MSE using an LBFGS gradient descent optimization solver. Data regularization is applied to the network inputs, the dataset is split into 75\% training and 25\% testing, and the network is trained performing multiple initializations in order to avoid local minima in the cost function.
The size of the hidden layer was determined experimentally by finding the minimum in the testing MSE, and it was set to 9 nodes.

\section{EXPERIMENTAL RESULTS} \label{sec:results}

\begin{table}[]
    \centering
    \begin{tabular}{|c||c|c|}
        \hline
        \textbf{Passenger Number} & \textbf{km} & \textbf{\%} \\
        \hline
        1 & 710 & 47.5 \\
        2 & 147 & 9.9 \\
        3 & 600 & 40.2 \\
        4 & 36  & 2.4 \\
        \hline
        \hline
        \textbf{Season} & \textbf{km} & \textbf{\%} \\
        \hline
        Summer & 476 & 31.9 \\
        Winter & 1017 & 68.1 \\
        \hline
    \end{tabular}
    \caption{Dataset distribution by passenger number and season}
    \label{tab: dataset_passengers}
\end{table}

\begin{figure}
    \centering
    \includegraphics{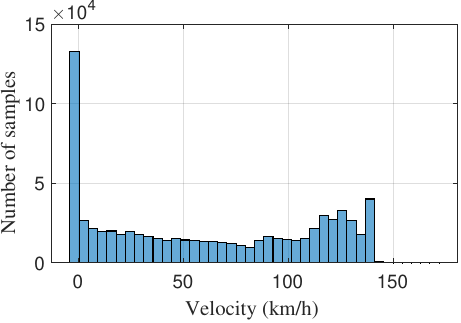}
    \caption{Dataset samples distribution by velocity}
    \label{fig: dataset_histogram}
\end{figure}    
\subsection{Dataset Description}

\begin{figure}
    \centering
    \includegraphics{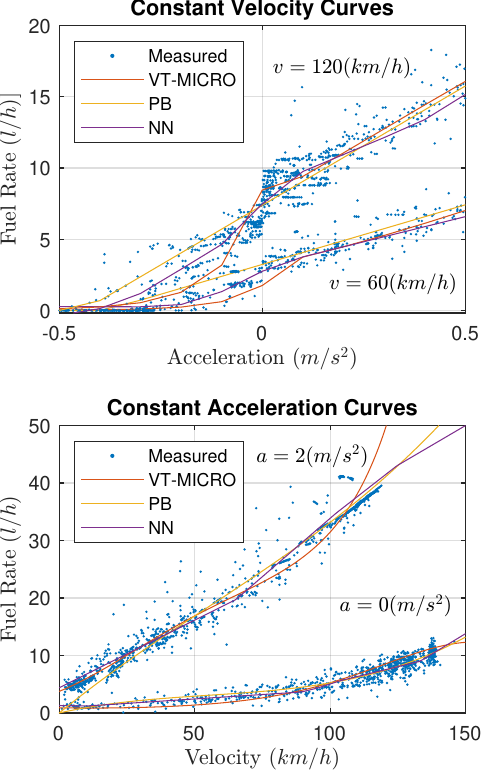}
    \caption{Level curves of the three models over experimental data. From the top: (a) constant velocity and (b) constant acceleration}
    \label{fig: level_curves}
\end{figure}

The experimental vehicle was driven over various roads in northern Italy for more than 20 hours over 1493  km, consuming 104.4 liters of diesel fuel. Table \ref{tab: dataset_passengers} describes the dataset composition in terms of the number of passengers and the season of the year, respectively. 
Fig. \ref{fig: dataset_histogram} shows the dataset distribution over speed, while Fig. \ref{fig: level_curves} shows the constant velocity and acceleration level curves and how the identified models fit the data.

Please note the large number of samples with $v=0$ in Fig. \ref{fig: dataset_histogram}: these are samples collected with the vehicle stopped, with either the engine idling or turned off manually, or by the start-and-stop system the car is provided with. 
Although they represent a large share of the total samples, the amount of fuel consumed in stopped vehicle conditions is just 3\% of the total in the dataset. We decided to exclude the stopped vehicle data from the algorithm design, calibration, and analysis and to study the stopped vehicle fuel consumption estimation problem separately in section \ref{stopped_car}.

\subsection{Instantaneous Fuel Consumption}


\begin{figure}
    \centering
    \includegraphics{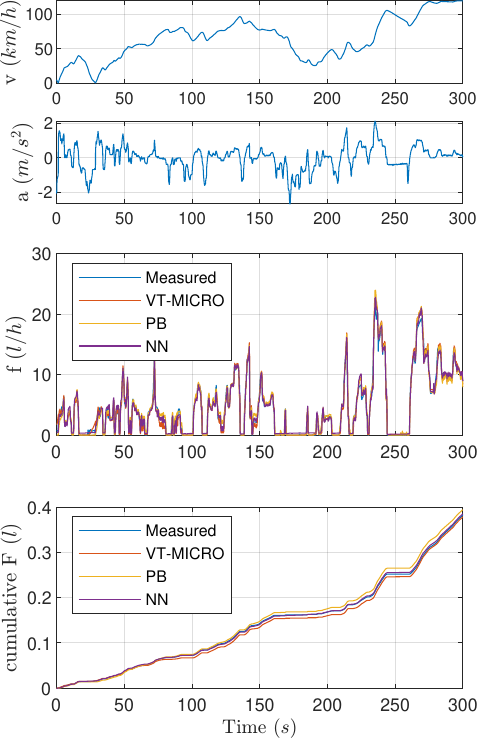}
    \caption{From the top: Measured velocity and acceleration; ground truth, and estimated instantaneous and integral fuel consumption of a mixed urban and highway driving experiment.}
    \label{fig: time_series}
\end{figure}

\begin{table}[]
    \begin{tabular}{|c|c|c|c|c|c|}
        \hline
        \multirow{2}{*}{\textbf{Model}}  & \multicolumn{2}{c|}{\textbf{Testing Error}} & \textbf{Integral} & \multicolumn{2}{c|}{\textbf{Error Over Tank}}  \\
        
        & \textbf{avg} (l/h) & \textbf{std} (l/h) &\textbf{Error} (\%l)  & \textbf{avg} (\%l)  & \textbf{std} (\%l) \\
        
        \hline
        
        VT-M    & +0.099             & 1.202     &-1.408           & -0.62             & 5.81 \\  
        
        PB          & -0.060             & 1.230   &
        \textbf{+0.376}    & \textbf{+0.03}    & \textbf{5.44} \\
        
        NN         & \textbf{-0.026}    & \textbf{1.028}     &-0.486           & -0.81             & 5.58 \\
        \hline
    \end{tabular}
    \caption{Experimental Results}
    \label{tab: results}
\end{table}

The evaluation of the instantaneous (sample-by-sample) fuel consumption estimation error provides a measure of how well the predicted values match the actual values while taking into account the scale of the data, and it penalizes large errors more than small errors.
The metric used to compare the three algorithms is, therefore, the Testing Error. 

This is a metric evaluating the three algorithms' capability of instantaneously predicting the fuel flow, and the results of the three considered fuel consumption virtual sensing methods are summarized in Table \ref{tab: results}. The Neural Network model is the best performer for this metric, with a 20\% standard deviation reduction w.r.t. the VT-MICRO model. 

Figure \ref{fig: time_series} shows the temporal performance of the three models on a single highway entrance experiment. This experiment was chosen in the testing dataset as it is representative of a wide range of velocity and acceleration.

\subsection{Integral Fuel Consumption}

In order to characterize the total energetic demand of a vehicle following a certain speed and acceleration profile over time, it can be useful to compare the three algorithms considering the error over the cumulative fuel consumption estimation over a certain time span.

Therefore, the second criterion employed to compare the three methods is the integral error, expressed in percentage as the difference between the integral of the estimated and measured fuel flow over the ground truth. This metric aims to optimize the quality of the estimation of the energy consumption given a certain speed and acceleration profile and should be less susceptible to measurement delays w.r.t. the Error metric.

As depicted in Fig. \ref{fig: time_series}, and confirmed in Table \ref{tab: results}, the PB and VT-MICRO models tend to over-estimate the fuel consumed, while the NN model tends to under-estimate the total consumed fuel. The PB model is the best performer according to this metric, as the module of the integral error is one order of magnitude lower than the VT-MICRO model, representing the state of the art.

\subsection{Error over Tank}

\begin{figure}
    \centering
    \includegraphics{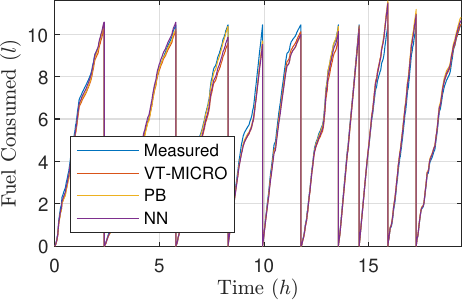}
    \caption{Computation of the Error Over Tank metric over the whole dataset. The integral of the fuel consumption is reset every 10.4 liters in order to provide different measurements}
    \label{fig: EOT}
\end{figure}

The last situation important metric used to evaluate the three algorithms refers to the idea of replicating the Integral Error metric over a series of subsequent trips in order for it to be more representative of the capability of the models to perform as a "virtual fuel gauge."

We can define the Error over Tank (EOT) metric: considering the whole dataset, it divides it into a set of "tanks" of total integral fuel consumed. Then, the integral error is computed for every "tank", as shown in Fig. \ref{fig: EOT} resulting in a distribution of errors, of whom the mean and standard deviation can be computed. 
This metric tries to be a more general version of the integral metric, as a model with (i.e.) the tendency to under-estimate at low speed and over-estimate at high speed would still have a good performance in a sufficiently balanced dataset. Splitting the dataset into a set of "tanks" allows us to compare the integral error over different conditions.

The results presented in Table \ref{tab: results} were obtained by dividing the whole dataset into 10 "tanks", each containing 10.4 liters of fuel. The PB model shows superior performance based on this metric: not only is it the only model with almost zero mean, but its standard deviation is the lowest of the considered trio. 
These results further confirm the tendency of the VT-MICRO and NN model to overestimate the total consumed fuel.

\subsection{Stopped Car Engine On Recognition and Fuel Consumption Estimation} \label{stopped_car}

\begin{figure}
    \centering
    \includegraphics{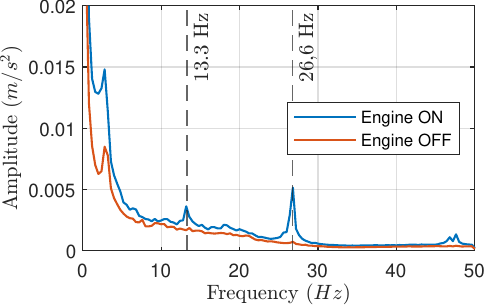}
    \caption{Average frequency spectrum content of acceleration data time series with the engine on and off respectively}
    \label{fig: FFT_eng_on_off}
\end{figure}

The aforementioned methodologies are valid for a moving vehicle. Instead, considering the case of a stopped vehicle, we have $v=0$ and $a=const.$ (due to the road slope) independently of whether the engine is on, with $f_{on}(t) = f_{idle}$ or off, with $f_{off}(t) = 0$.

Even if the $v=0$ condition is responsible for only 3\% of the total fuel consumed in the dataset, this percentage increases up to 20\% if we consider only urban driving conditions ($v<50$ $km/h$). A static model with only instantaneous $v$ and $a$ measurements as inputs is not able to distinguish between the two situations and will inevitably converge to an estimated fuel value in the interval $[0,f_{idle}]$.

A possible approach to identify the engine's on and off conditions is to perform a frequency analysis over the raw acceleration data, looking for the harmonics of the engine rotation speed, which should be present only in the data with the engine on. 
To perform this analysis, it is necessary to extract and process the raw IMU longitudinal acceleration measurements, sampled at 100 Hz, for every continuous segment of $v=0$ velocity data. Then, the frequency spectra are averaged over all the segments with the engine on and off, respectively.

The results of this analysis are presented in Fig. \ref{fig: FFT_eng_on_off}, showing that the average spectrum of the acceleration with the engine on presents two resonant peaks at 13.3 and 26.6 Hz, corresponding to the first and second harmonics of the engine idle rotational speed (800 revolutions per minute).
A simple classifier using a threshold over the second peak frequency to determine whether the engine is on or not was able to achieve 92\% for both the True Positive Rate (TPR) and Positive Predicted Value (PPV) metrics. 

However, this methodology suffers from a lack of generality, as it proved to be highly susceptible to the specific considered vehicle and to the IMU mounting conditions, especially concerning the sensor placement on the vehicle and choice of mounting fasteners.

\addtolength{\textheight}{-2cm}   

\section{FINAL REMARKS} \label{conclusions}

This paper presents a comparison of three algorithms for estimating fuel consumption from an external, removable GNSS and IMU sensor unit mounted inside the vehicle. Two new algorithms, PB and NN, were compared with the best state-of-the-art model. The dataset used for the experimental validation was composed of GNSS velocity and IMU acceleration data collected during several trips in different seasons and with varying numbers of passengers.

The three algorithms were evaluated based on three metrics: Testing Error, Integral Error, and Error over Tank. The experimental results showed that the Neural Network model was the best performer in terms of Testing Error, while the PB model was the best performer in terms of Integral Error and Error over Tank.
In any case, both proposed algorithms, Neural Network and PB, outperform the state of the art by one order of magnitude in almost every metric.

The paper also discussed the problem of estimating fuel consumption for stopped vehicles with the engine on, providing a separate analysis of this problem using frequency spectrum analysis techniques. Although the results looked promising over this dataset, the proposed methodology lacks generality but still represents a promising starting point for future works.

While the current study focuses on the fuel consumption estimation for a diesel vehicle, the methodology and algorithms presented in this paper can be generalized to other types of vehicles, including gasoline but also hydrogen or electric vehicles.

However, there are some differences in the characteristics of these types of vehicles that will need to be taken into account. For example, in electric vehicles, the regenerative braking system can produce negative charge consumption values.
Future developments of this research may include testing these models over different classes of vehicles to see if the method can be applied more broadly. Additionally, testing how a model calibrated on a specific vehicle performs on a different vehicle class could provide valuable insights into the three algorithms' scalability. 

These advancements in vehicle emission modeling can aid in the development of more efficient and sustainable transportation solutions and traffic policies, benefiting both  environment and public health.

\section*{ACKNOWLEDGMENTS}
The authors gratefully acknowledge the financial and technical support of UnipolTech S.p.A.

\IEEEpeerreviewmaketitle

\bibliographystyle{IEEEtran}
\bibliography{IEEEabrv,ConsumiUnipol}

\end{document}